\documentclass[final]{cvpr}

\usepackage{times}
\usepackage{epsfig}
\usepackage{graphicx}
\usepackage{amsmath}
\usepackage{amssymb}
\usepackage[toc,page]{appendix}
\DeclareMathOperator*{\minimize}{minimize}

\usepackage[pagebackref=true,breaklinks=true,colorlinks,bookmarks=false]{hyperref}

\def\cvprPaperID{****} 
\def\confYear{CVPR 2021}

\begin{document}

\title{Confounding Tradeoffs for Neural Network Quantization}

\author{Sahaj Garg\\
Luminous Computing\\
Mountain View, CA\\
{\tt\small sahajgarg@gmail.com}
\and
Anirudh Jain\\
Luminous Computing\\
Mountain View, CA\\
{\tt\small anirudh1097@gmail.com}
\and
Joe Lou\thanks{Work completed while the author was with Luminous Computing.}\\
Facebook\\
Menlo Park, CA\\
{\tt\small zlou@fb.com}
\and
Mitchell Nahmias\\
Luminous Computing\\
Mountain View, CA\\
{\tt\small mitch@luminouscomputing.com}
}

\maketitle

\newcommand{\bfx}{\mathbf{x}}
\newcommand{\bfw}{\mathbf{w}}
\newcommand{\bfW}{\mathbf{W}}
\newcommand{\bfX}{\mathbf{X}}
\newcommand{\bfA}{\mathbf{A}}
\newcommand{\bfa}{\mathbf{a}}
\newcommand{\bfE}{\mathbf{E}}
\newcommand{\bftheta}{\mathbf{\theta}}
\newcommand{\xreal}{\bfx}
\newcommand{\wreal}{\bfw}
\newcommand{\xq}{\bfx^{(q)}}
\newcommand{\wq}{\bfw^{(q)}}
\newcommand{\yq}{y^{(q)}}
\newcommand{\zr}{z}
\newcommand{\R}{\mathbb{R}}
\newcommand{\N}{\mathcal{N}}
\newcommand{\ytilde}{\tilde{y}}
\newcommand{\ytq}{\tilde{y}^{(q)}}
\newcommand{\ztr}{\tilde{z}}

\newcommand{\bfB}{\mathbf{B}}

\begin{abstract}
Many neural network quantization techniques have been developed to decrease the computational and memory footprint of deep learning.
However, these methods are evaluated subject to confounding tradeoffs that may affect inference acceleration
or resource complexity in exchange for higher accuracy. In this work, we articulate a variety of tradeoffs whose impact is often overlooked 
and empirically analyze their impact on uniform and mixed-precision post-training quantization, finding that these confounding tradeoffs may have a larger impact on quantized network accuracy than the actual quantization methods themselves.
Because these tradeoffs constrain the attainable hardware acceleration for different use-cases, we encourage researchers to explicitly report these design choices through the structure of ``quantization cards.'' We expect quantization cards to help researchers compare methods more effectively and engineers determine the applicability of quantization techniques for their hardware. 
\end{abstract}

\section{Introduction}
The recent success of deep learning has come at the expense of larger models, which have greater computational cost and memory footprint, making the deployment of neural network inference challenging. A common approach for improving the  performance of neural network inference is quantization, which represents floating point values in neural networks as low-bit integers \cite{Jacob:quantization, Krishnamoorthi:tf-whitepaper}. By utilizing hardware support for low precision arithmetic, quantization enables higher throughput, decreased memory traffic, and lower storage requirements.

Many techniques have been developed to enable quantized inference in a variety of settings. 
If adequate resources are available, models can be retrained for multiple epochs while simulating the effects of quantization \cite{Krishnamoorthi:tf-whitepaper, esser:lsq}, but in other cases, no data may be available for re-training 
\cite{cai:zeroq, xu:generative, nagel:dfq}. Depending on hardware support and available resources, quantization strategies 
use mixed precision
\cite{Uhlich:learn_bw, HAWQ, HAWQv2, haq, dnas, banner:fourbit}, optimize for more effective quantization grids and clipping ranges \cite{banner:fourbit, Choukroun:omse, zhao2019ocs}, or define mappings from weights to quantized values that are more effective than rounding \cite{wang:bitsplit, nagel:adaround, hubara:adaquant}. 

A key challenge in the quantization literature is attributing improvements in accuracy to methods because different methods are evaluated subject to confounding tradeoffs.
These confounding tradeoffs represent design decisions that trade off inference acceleration
or resource complexity for improved accuracy. In Section \ref{sec:tradeoffs}, we articulate a variety of these confounding tradeoffs, including the quantization method (signed vs. unsigned, per-tensor vs. per-channel), which operations are quantized (e.g. residual connections), mixed-precision constraints, the amount of data and compute required by the method, and the baseline accuracy of the floating point model. 

To address this problem, we propose that novel quantization research discuss these tradeoffs through ``quantization cards,'' described in Section \ref{sec:quant_card}. Quantization cards are a structure for reporting design decisions, analogous to datasheets for hardware, and more recently, datasheets for datasets \cite{datasheets:gebru} or model cards for machine learning models \cite{mitchell:modelcards}. Such a structure is important for reporting quantization design decisions because unlike other hyperparameters, these design decisions affect the attainable inference acceleration.
We propose quantization cards instead of establishing a benchmark because the set of tradeoffs is rapidly growing, such as using integer-only operations excluding division \cite{yao2020hawqv3} or generating synthetic data \cite{cai:zeroq, xu:generative}.

We expect quantization cards to be useful for two groups of people. First, researchers of quantization techniques will be able to more accurately compare across methods and replicate prior results. Second, engineers deploying quantized models will be able to assess quantization cards, determine whether the tradeoffs made by the work are applicable for their hardware and application setting, and adopt or combine strategies accordingly. Without quantization cards, these choices are often scattered throughout different parts of the text or even overlooked, making it challenging for users to determine how the method can be used.

We evaluate the impacts of these tradeoffs on quantized model accuracy in Section \ref{sec:experiments} for uniform quantization and post-training mixed-precision quantization. 
These experiments demonstrate that changes in confounding tradeoffs can affect accuracy at low bit precision by multiple percentage points, emphasizing the need for quantization cards and unified reporting of design decisions.

\section{Confounding Tradeoffs}
\label{sec:tradeoffs}
Quantization methods are evaluated under a wide array of design choices, each of which trades off inference acceleration or resource complexity. Here, we discuss how those tradeoffs manifest and their impacts on accuracy.

We specifically discuss tradeoffs for uniform quantization, the most common method for quantizing floating point values to low precision \cite{Jacob:quantization}. In uniform quantization, floating point values $\bfx$ in some range $[\bfx_{min}, \bfx_{max}]$ are uniformly mapped to $B$ bit integers by scaling, translating, and rounding the values. Mathematically, this is $\bfx_q = \textrm{round}\left(\frac{\bfx}{\Delta} + b\right)$, where the scale $\Delta = \frac{\bfx_{max} - \bfx_{min}}{2^B - 1}$, and the bias $b$ depends on whether the integers are signed or unsigned. The quantization parameters can be determined by calibration of the range on some subset of data \cite{Jacob:quantization} or other strategies such as minimizing the mean squared quantization error \cite{Choukroun:omse}. On the other hand, there are various approaches to non-uniform quantization, which may or may not be compatible with bit-operations in hardware \cite{zhang:lqnet, miyashita:logarithmic}. We restrict our focus in this paper to confounding tradeoffs for the more commonly deployed uniform quantization. 

\subsection{Quantization Method}
\paragraph{Symmetric vs. Asymmetric Quantization}
Uniform quantization is typically applied via symmetric quantization to signed integers or asymmetric quantization to unsigned integers \cite{Krishnamoorthi:tf-whitepaper}. Symmetric quantization enforces that $\bfx_{min} = -\bfx_{max}$, so the bias of the quantization operation is zero, and the floating point value zero is mapped to the quantized integer zero. On the other hand, asymmetric quantization applies to an arbitrary data range, so it can more effectively capture $\bfx$ that is not symmetrically distributed about zero, but consequently will have nonzero bias $b$. The ability to capture an asymmetric data range is especially important for activations after the ReLU nonlinearity, which discards the sign bit for symmetric quantization. As a result, asymmetric quantizers are expected to outperform symmetric quantizers in accuracy. 

However, asymmetric quantization may decrease inference acceleration. In order to evaluate a real valued matrix multiplication using quantized inputs,  it is necessary to revert the result by multiplying the bias of $\bfx$ by the sum of the weights in the dot product, and the bias of $\bfW$ by the sum of the inputs \cite{Krishnamoorthi:tf-whitepaper}. While these sums can be computed at lower complexity than the matrix multiplication, they present a nontrivial and potentially undesirable overhead. Because the sum of $\bfx$ cannot be computed ahead of time, asymmetric quantization of weights presents the larger overhead. 

\paragraph{Per-channel vs. Per-Tensor Quantization}
A frequently discussed tradeoff is between using per-tensor quantization, which quantizes all values in a matrix using the same minimum and maximum, and per-channel quantization, which uses a different range for each channel of a weight matrix or activation tensor \cite{wang:bitsplit, nagel:adaround}. Using per-channel quantization improves accuracy at the expense of greater complexity from applying requantization to each channel independently. Moreover, it may be challenging to achieve inference acceleration for dot products if different activations in the same dot product are quantized to different scales.

\subsection{Quantized Operations}
\begin{figure}[]
    \centering
    \includegraphics[width=\linewidth]{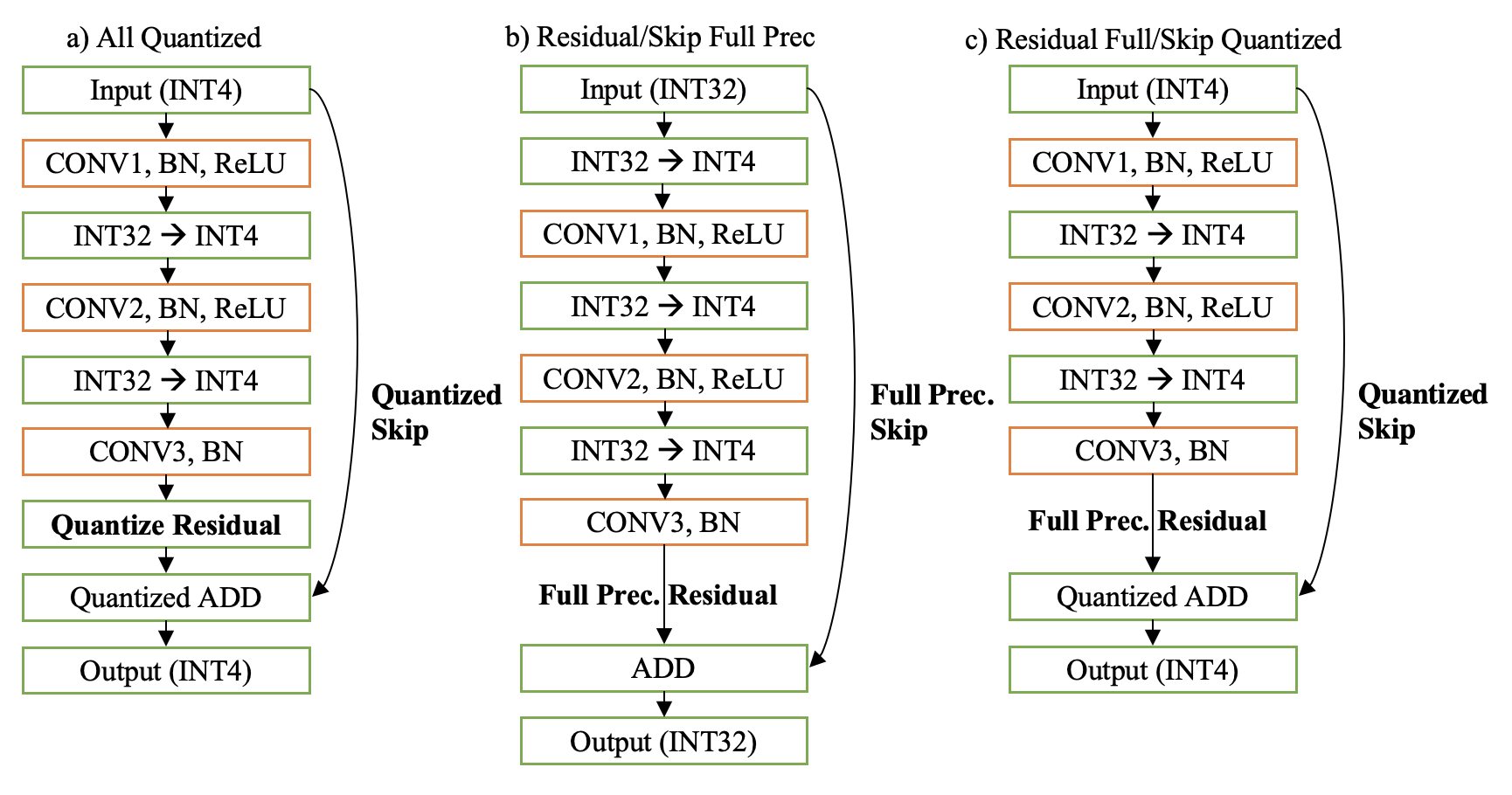}
    \caption{Different approaches to quantizing residual connections.}
    \label{fig:skip}
\end{figure}

\paragraph{Elementwise Operations and Residual Connections}
While most quantization methods guarantee that the inputs to every matrix multiplication are quantized, many take different approaches for the quantization of other elementwise operations, such as residual connections. The most common method, and the one implemented in TensorFlow and PyTorch, assumes that the results of every operation are requantized, implying that the inputs to a residual addition are low bit integers, and the output of the elementwise addition is requantized \cite{Krishnamoorthi:tf-whitepaper}. However, because elementwise operations have a relatively smaller computational complexity than matrix multiplication, they can be executed at higher precision before being requantized \cite{Liu:bireal, choi:pact}. Different methods for quantizing residual connections are illustrated in Figure \ref{fig:skip}. When skip connections are evaluated at high precision, this creates high precision activation propagation through the entire network \cite{park:highway}, improving accuracy when using low precision activations. 

Full precision elementwise operations, while tractable in many scenarios, may lead to inefficiencies for certain hardware architectures. When memory pressure and bandwidth is a constraint, high precision intermediate feature maps may not fit in cache, and thus must be stored and fetched from main memory. The increased memory pressure may increase inference latency. 

Other elementwise operations present similar confounding tradeoffs. For computer vision models, average pooling may be performed at high precision before being requantized, or may be performed directly with low precision integer arithmetic, which leads to information loss \cite{yao2020hawqv3}.  
Quantization of average pooling may be overlooked when quantization is simulated with floating point arithmetic.  
On the other hand, using integer-only arithmetic avoids the cost of casting integers to and from floating point values or performing expensive division for elementwise operations, and improves performance  \cite{yao2020hawqv3}.
For natural language processing, transformer models such as BERT contain many elementwise operations, including adding token embeddings to positional embeddings, intermediate nonlinearities such as the softmax, and normalization layers such as LayerNorm that cannot be fused. These operations are often computed in FP32 \cite{zafrir:q8bert, zhang-etal-2020-ternarybert, bai2020binarybert}, but can be computed more efficiently via integer-only approximations \cite{kim:ibert}. 

\paragraph{Excluding First and Last Layer}
Many works quantize the first and last layer of computer vision models to higher precision, e.g. 8 bits, because they are especially precision sensitive \cite{banner:fourbit}. However, because the last layer of Resnet50 is extremely large, at 3 bits, quantizing the first and last layer at 8 bit precision requires 13\% more memory capacity. This difference in compression ratio may improve accuracy at ultra-low precision, so the compression ratio must account for the size of the first and last layer. 

\subsection{Mixed Precision}
Different layers of neural networks are tolerant to different degrees of precision, and uniformly quantizing all layers of a neural network to the same low precision leads to accuracy degradation \cite{HAWQv2}. Mixed-precision quantization uses varying precision for different layers \cite{Uhlich:learn_bw, HAWQ, HAWQv2, haq, lin:sqnr, dnas, nahshan:lossaware} or channels \cite{banner:fourbit, yang2020fracbits} of neural networks to decrease the average precision required by neural networks. 

Mixed precision allows for different settings for the number of bits. For example, some works restrict the bitwidth for quantized operations to $\{2, 4, 8\}$ bits \cite{cai:zeroq}, whereas other works allow any integer bitwidth \cite{Uhlich:learn_bw, banner:fourbit}. While arithmetic on 3-bit integers may not be directly supported in hardware, and operations must be computed by casting these integers to 4-bits, such approaches can save memory traffic if values are efficiently packed. Describing the choice of bitwidth is critical because one mixed-precision method may outperform another solely because it has a larger selection of bit precision settings. 

Moreover, the activation bitwidth utilized by a mixed-precision network may be constrained by different desiderata. If memory capacity is the limiting constraint for acceleration, then activation bitwidths can be constrained solely by the maximum feature map size \cite{Uhlich:learn_bw}. However, if memory pressure or computational throughput is the system bottleneck, mixed precision methods may be constrained by the average feature map size. The choice of constraint will impact accuracy.

\subsection{Resource Complexity} 

Most approaches to quantization are split into three categories, zero-shot quantization, post-training quantization, and quantization-aware training, each of which require different amounts of data and computation. Post-training and zero-shot quantization are typically applied in scenarios where access to data may be limited due to e.g. privacy concerns \cite{cai:zeroq, banner:fourbit}. However, we observe substantial variation within each of these categories in their data and computational complexity. 
\begin{table*}[t!]
\centering
\begin{tabular}{|p{0.2\linewidth}|p{0.75\linewidth}|}
 \hline
         &  Design Tradeoffs \\  \hline
Quantization Method &  Uniform/non-uniform $\vert$ Symmetric/asymmetric  $\vert$ Per-channel/Per-tensor $\vert$ Dynamic/Static \\ \hline
Quantized Operations &  Skip connection $\vert$ Residual connection $\vert$ First/last layer  $\vert$ Pooling \\ \hline
Mixed Precision &  Allowed bitwidths $\vert$ Average or maximum feature map size   \\ \hline
Resource Complexity &  Amount of data $\vert$ Type of data (synthetic? labeled?) $\vert$ Time (fraction 1 epoch train)   \\ \hline
Pretrained Model &  Baseline accuracy $\vert$ Quantized accuracy  \\ \hline
\end{tabular}
\caption{Quantization Card, with suggested design tradeoffs}
\label{tab:modelcard}
\end{table*}       

\paragraph{Data Complexity}
The data complexity of different methods may be viewed on a sliding scale. Approaches for post-training quantization assume different amounts and types of data, ranging from 256 \cite{Liu:multipoint} to 1024 examples \cite{nagel:adaround}, and sometimes using one example per class \cite{hubara:adaquant}. In addition, some methods only require unlabeled calibration data, whereas others (such as the one presented in this paper) require access to labeled data, which may be prohibitive in privacy constrained settings. However, it is possible to generate synthetic labeled data from pretrained models and to use this data for quantization \cite{cai:zeroq, xu:generative}. This may lift the stringent ${<}$1000 sample requirement for post-training quantization, and instead enable post-training quantization with the only requirement that it be performed rapidly.

\paragraph{Computational Complexity}
Similarly, the computational complexity of different methods ought to be viewed on a sliding scale. 
There is a growing body of work that uses different types of optimization or even backpropagation for post-training quantization, but that can complete within 10 minutes on a GPU \cite{nagel:adaround}. Moreover, methods that perform quantization-aware training for a single epoch versus fifty epochs are different in their applicability; the former method may be usable to deploy the same pretrained model to multiple different resource constrained environments \cite{Cai2020Once-for-All}. 

\subsection{Pretrained Model Accuracy}
When evaluating post-training quantization for Resnet50, different works use floating point baseline models ranging in validation accuracy from the torchvision model with 76\% accuracy \cite{banner:fourbit, Choukroun:omse, Liu:multipoint} to the pytorchcv model with 77.7\% accuracy \cite{cai:zeroq, hubara:adaquant}. However, most works compare their results to those reported by previous papers, so post-training quantization works that use a model with 1.7\% better floating point accuracy will by default outperform prior work when the accuracy degradation is small. Thus, studies should note baseline accuracy and relative accuracy degradation when comparing across works. They should also default to baselines with the highest accuracy, as suggested in  \cite{yao2020hawqv3}.

\section{Quantization Cards}
\label{sec:quant_card}

We propose that researchers of quantization methods describe the design tradeoffs used when evaluating their method through ``quantization cards.''  Quantization cards are broken down into the five categories from Section \ref{sec:tradeoffs} and explain the tradeoffs made for each of these categories. The categories and specific tradeoffs within each category will grow over time as new tradeoffs are developed. 
In Table \ref{tab:modelcard} we show the generic questions that should be answered by a quantization card, and in Section \ref{sec:discussion}, we provide an illustrative example. When multiple design settings are used for different experiments in a single work, a single quantization card for the strongest method in the paper with comments on deviations from those settings is sufficient. 

Quantization cards may be useful to both researchers of quantization methods and engineers deploying quantized models. First, quantization cards identify often overlooked tradeoffs that impact accuracy by ${>}1\%$, which can be larger than the difference between state-of-the-art techniques. This helps distinguish improvements of novel methods from those of design decisions and facilitates better comparisons across methods. Second, they improve reproducibility by guaranteeing that all confounding tradeoffs are described. Third, by reporting these tradeoffs in a single location, engineers can at a glance determine whether these tradeoffs can be made by their system, or whether the technique must be adapted. Engineers require this information to be clearly presented because it affects the attainable inference acceleration and the use cases in which methods can be deployed, unlike other hyperparameters, which only affect reproducibility of results.


We suggest the use of quantization cards instead of creating a new benchmark because the set of design tradeoffs is rapidly evolving. For example, some methods restrict operations to integer-only arithmetic excluding division \cite{yao2020hawqv3}, approximate nonlinear operations such as softmax using integer-only arithmetic for NLP applications \cite{kim:ibert}, or generate synthetic data, which may alleviate stringent data constraints for post-training quantization \cite{xu:generative, cai:zeroq}. Because these tradeoffs involve novel ways of implementing quantization, 
they cannot be captured by a fixed benchmark.
Instead, quantization cards are modeled after proposals for describing characteristics and use cases of models \cite{mitchell:modelcards} or datasets \cite{datasheets:gebru, holland:nutrition, bender-friedman-2018-data, factsheets:arnold} to address issues related to fairness and bias in machine learning. They also resemble guidelines for transparent reporting of prediction models in health \cite{collins:tripod}.

\section{Experiments}
\label{sec:experiments}
\subsection{Method}

We design experiments for each of the design tradeoffs in Section \ref{sec:tradeoffs} and evaluate results on Resnet50 \cite{he:resnet} on the Imagenet dataset \cite{deng:imagenet}. Experiments assume uniform precision unless otherwise noted, keeping the first and last layer at 8 bit precision. For experiments using uniform quantization at low activation bitwidth, we use percentile clipping of activations at the 99.99th percentile \cite{li:percentileclip, mckinstry:percentile}.

For experiments on mixed-precision quantization, we propose a method to stress test the minimum attainable bit precision of different computations. We do so by leveraging the method in \cite{Uhlich:learn_bw} to learn the optimal bitwidth for weights and activations in each layer
for a pretrained model. This amounts to solving an optimization problem over the bitwidths of each layer to minimize the task loss with an added penalty for the average bitwidth of weights and activations. Experiments using learned mixed-precision quantization are repeated with 5 random seeds, and mean and standard deviation of results is plotted. Uniform quantization has lower variance because it does not use learned parameters, so we evaluate results with a single seed. The optimization problem is formally described in  Appendix \ref{app:method}. 

Details of each experimental configuration are described in Appendix \ref{app:details}, and code is open sourced at \url{https://github.com/sahajgarg/low_precision_nn}.

\subsection{Results}

\paragraph{Observation 1: Asymmetric quantizers require fewer bits than symmetric quantizers for equivalent accuracy.} $\quad$

\begin{figure}
    \centering
    \includegraphics[width=\linewidth]{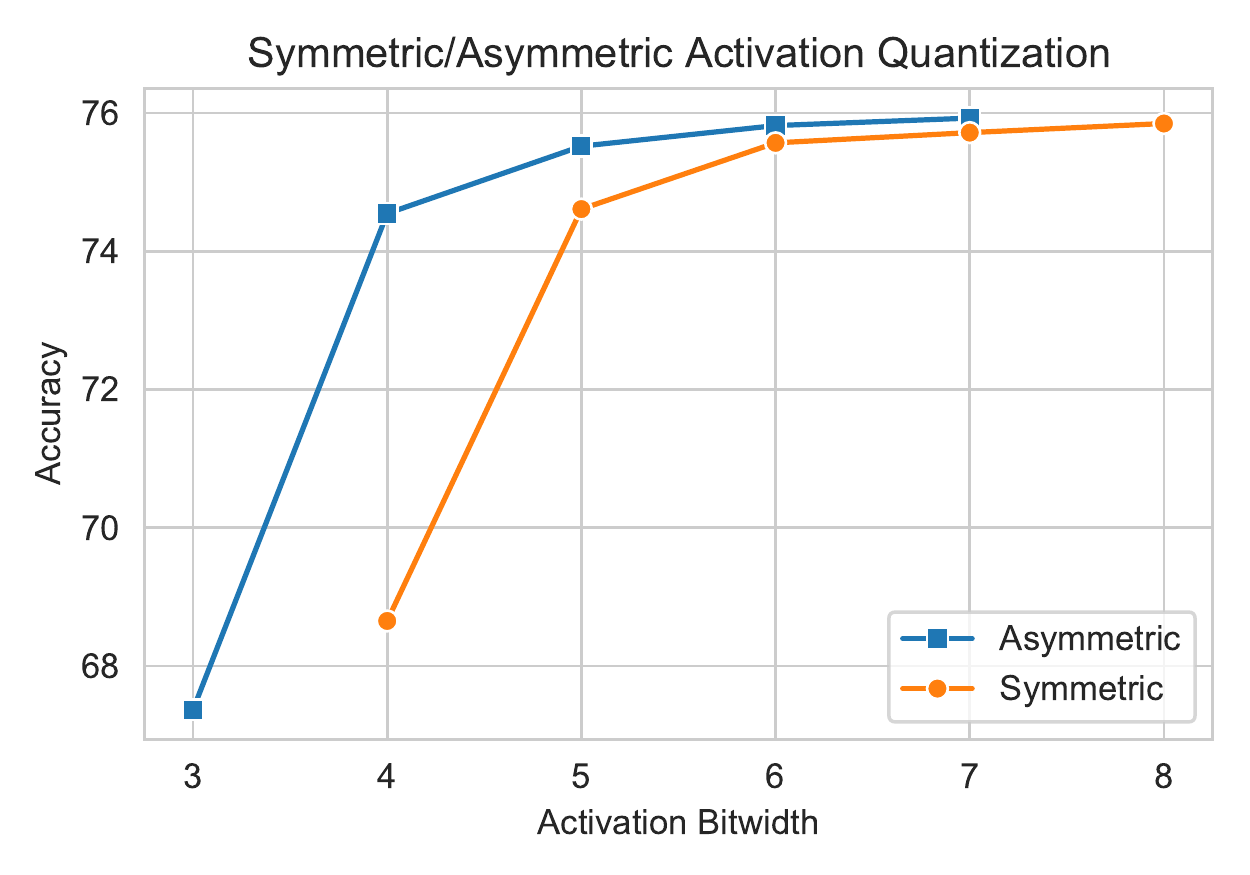}
    \caption{Symmetric activation quantization requires one more bit.}
    \label{fig:unif-symmetric}
\end{figure}

We test the impact of asymmetric quantizers on activation quantization, where we expect that applying symmetric quantization to unsigned activations will require one additional bit due to the discarded sign bit.
We quantize all operations except for residual connections in Resnet50, which corresponds to quantizing all operations after the output of a ReLU nonlinearity. We report results in Figure \ref{fig:unif-symmetric}. Accuracy does not precisely match up between 4 asymmetric bits and 5 symmetric bits because both must include zero in their representation. 

\begin{figure}
    \centering
    \includegraphics[width=\linewidth]{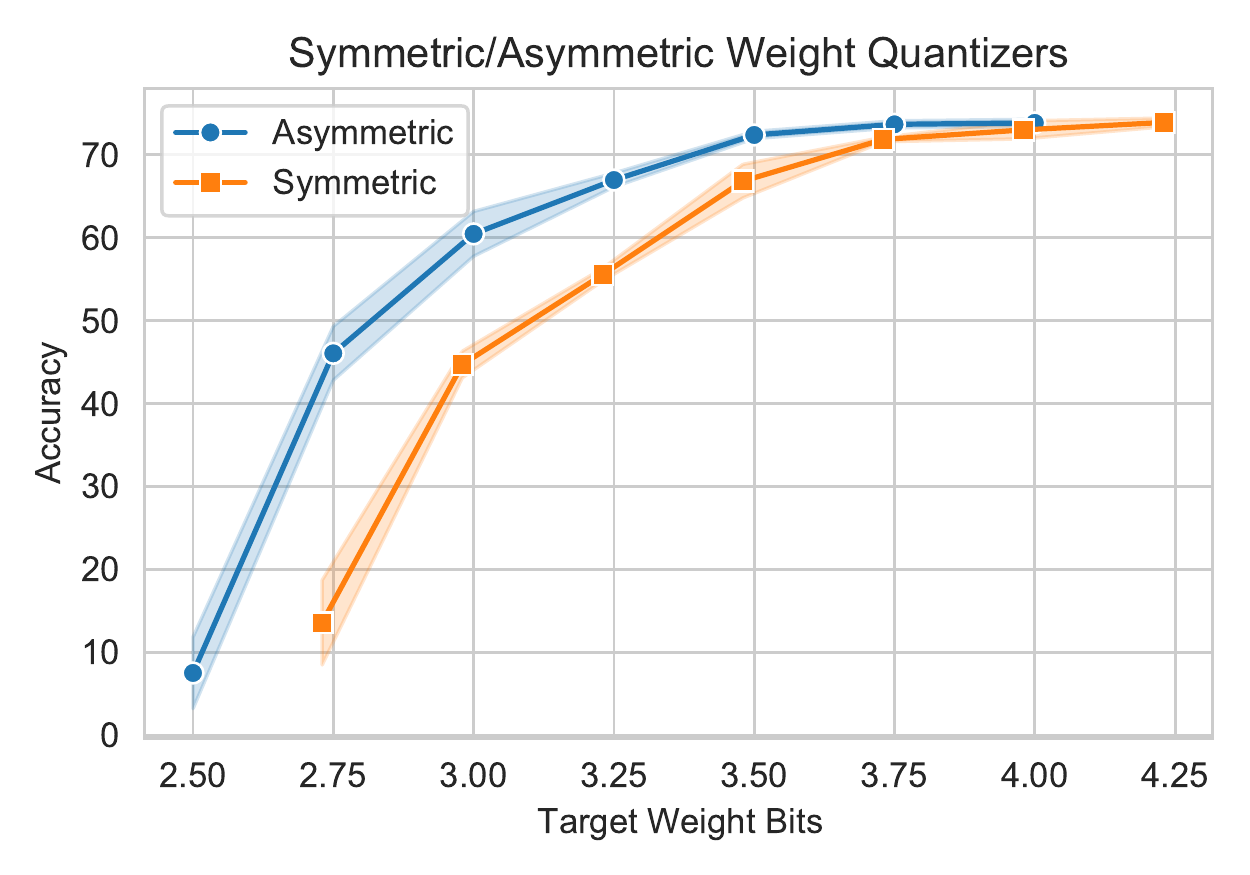}
    \caption{Asymmetric weight quantizers can use approximately 0.23 fewer bits on average while obtaining similar accuracy for Resnet50.}
    \label{fig:weight-symmetric}
\end{figure}

The impact of asymmetric quantizers on weight quantization is smaller since weights are typically signed, although their distributions may be skewed. The number of quantization bins necessary to represent weights with equal fidelity for symmetric and asymmetric quantizers can be measured by the ratio of their ranges, i.e. $\bfW_{max} - \bfW_{min}$ versus $2 \left(\operatorname{max}\left(\bfW_{max}, -\bfW_{min}\right)\right)$. If the number of bins could be modulated (as opposed to the coarser-grained number of bits), using asymmetric weights would require on average 0.23 fewer bits for Resnet50, or approximately 730 KB. To test the impacts of asymmetric weight quantization when the bitwidths must be discrete, we evaluate mixed-precision weight quantization for asymmetric quantizers and symmetric quantizers that use an additional 0.23 bits on average, reported in Figure \ref{fig:weight-symmetric}. At low bitwidths, asymmetric quantization improves accuracy by multiple percentage points.

\paragraph{Observation 2: Per-channel quantization outperforms per-tensor quantization. } $\quad$

\begin{figure}
    \centering
    \includegraphics[width=\linewidth]{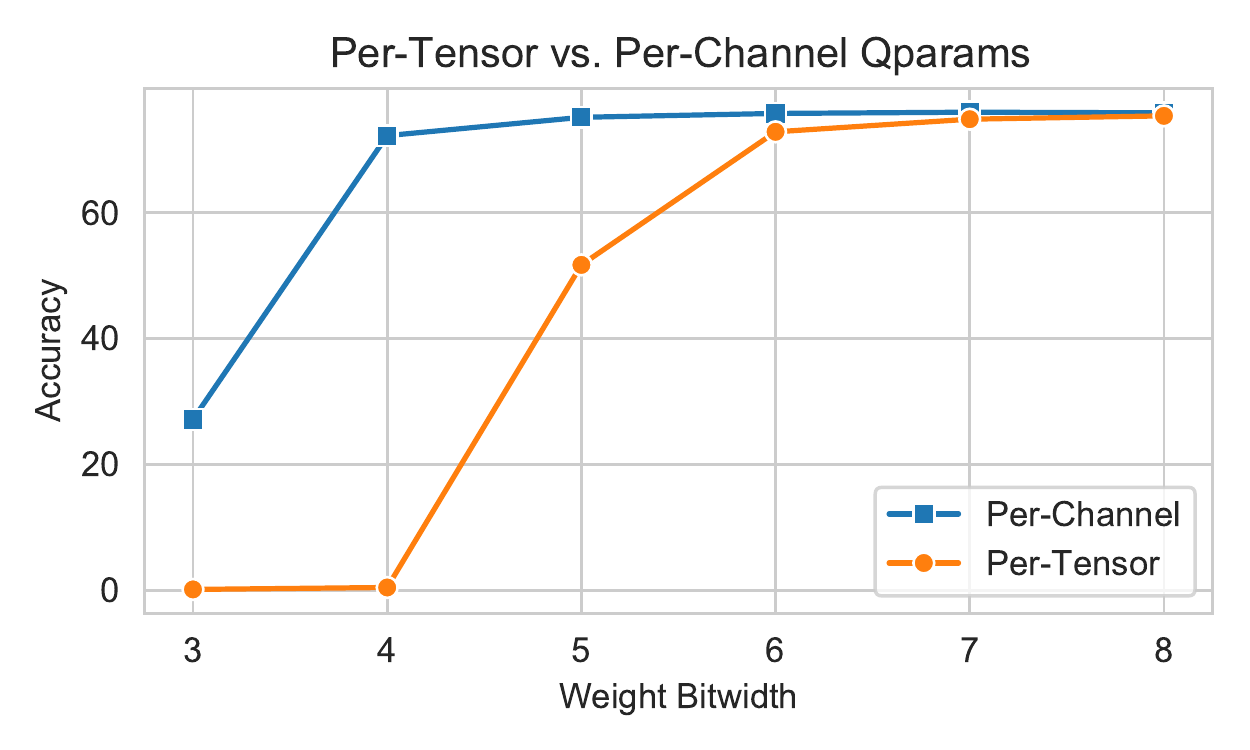}
    \caption{Per channel quantization parameters and per channel bitwidth allocation improve quantized model accuracy.}
    \label{fig:channel}
\end{figure}

We confirm this well-known observation for weight quantizers in Figure \ref{fig:channel}. 

\paragraph{Observation 3: If residual and skip connections are not quantized, methods can operate at much lower average activation precision.} $\quad$

\begin{figure}
    \centering
    \includegraphics[width=\linewidth]{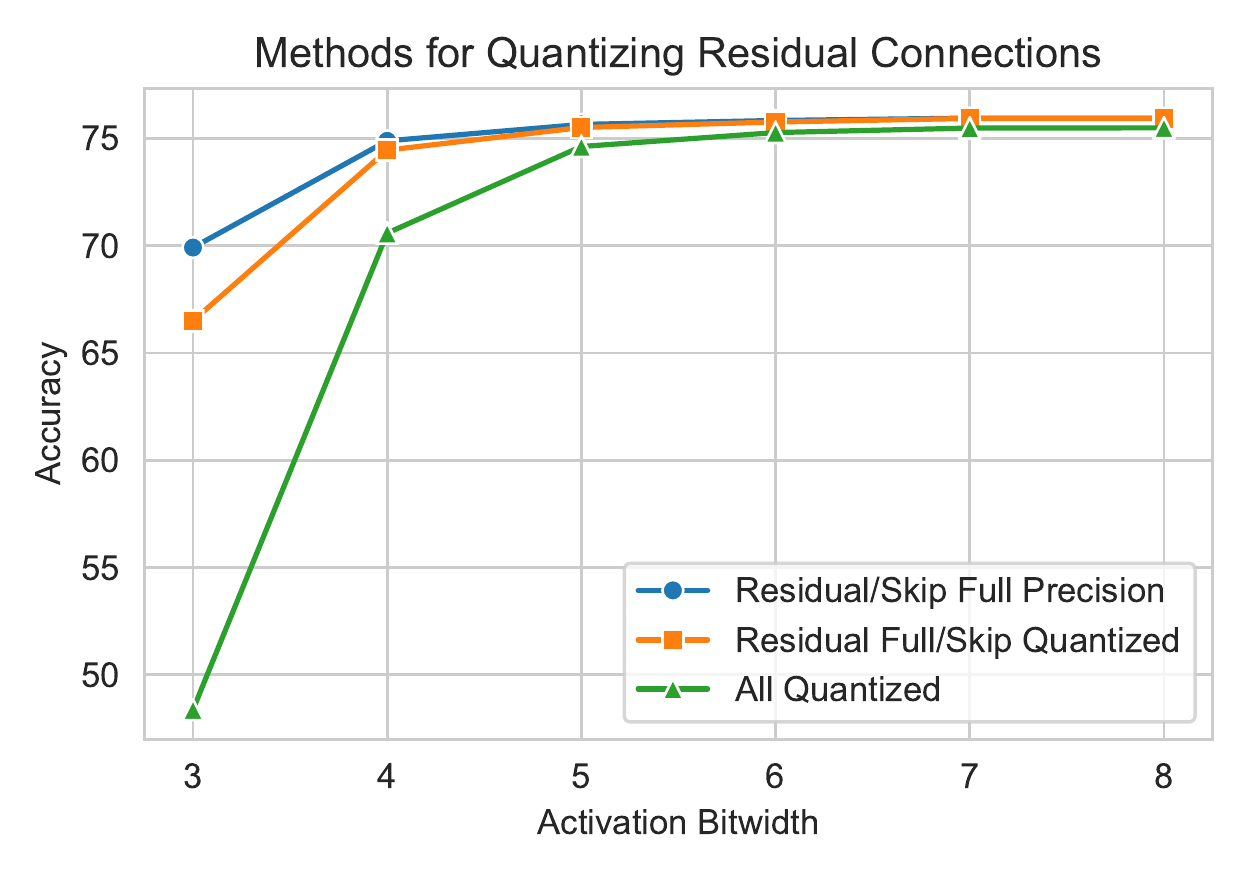}
    \caption{Methods that do not quantize skip and residual connections may be able to quantize models to 2-3 bit activations, whereas methods that quantize all operations require more bits to retain accuracy.}
    \label{fig:highway_results}
\end{figure}

We compare the three methods of quantizing residual and skip connections from Figure \ref{fig:skip} on their accuracy at low activation precision. 
Results in Figure \ref{fig:highway_results} demonstrate that methods such as \cite{wang:bitsplit} that attain extremely high accuracy with 3 bit activations may be able to do so in part because they do not quantize residual and skip connections. 

\paragraph{Observation 4: Ignoring the first and last layer affects the compression ratio, and consequently accuracy when subject to mixed-precision compression targets.} $\quad$

\begin{table}[]
\centering
\begin{tabular}{|p{0.08\textwidth}|p{0.06\textwidth}|p{0.09\textwidth}|p{0.035\textwidth}|l|}
\hline
Model Size & Weight Bits & Total Feat Map  & Act Bits  & Accuracy \\ \hline
12.8 MB & 4        & 8.3 MB & 4          & 71.6 $\pm$ 0.3    \\ \hline
13.8 MB & 4.32     & 8.7 MB & 4.19      & 72.4 $\pm$ 0.5 \\ \hline
12.8 MB & 4        & 16.6 MB & 8  &   74.0 $\pm$ 1.2   \\ \hline
13.8 MB & 4.32     & 16.6 MB & 8      & 74.8 $\pm$ 0.1   \\ \hline
\end{tabular} 
\caption{Ignoring the first and last layer affects the compression ratio.}
\label{tab:first+last}
\end{table}

The amount of model compression is affected by whether the first and last layer are quantized. We train mixed-precision bitwidth allocations that quantize to an average of 4 bits for weights and activations over all layers and compare against mixed-precision models that keep the first and last layer at 8 bits while quantizing all other layers to an average of 4 bits. While the difference in compression might appear small, results in Table \ref{tab:first+last} show that it can affect accuracy by almost $1\%$. 

\paragraph{Observation 5: The set of allowed bitwidths for mixed-precision quantization affects the ability of mixed precision to increase accuracy.} $\quad$ 

\begin{figure}
    \centering
    \includegraphics[width=\linewidth]{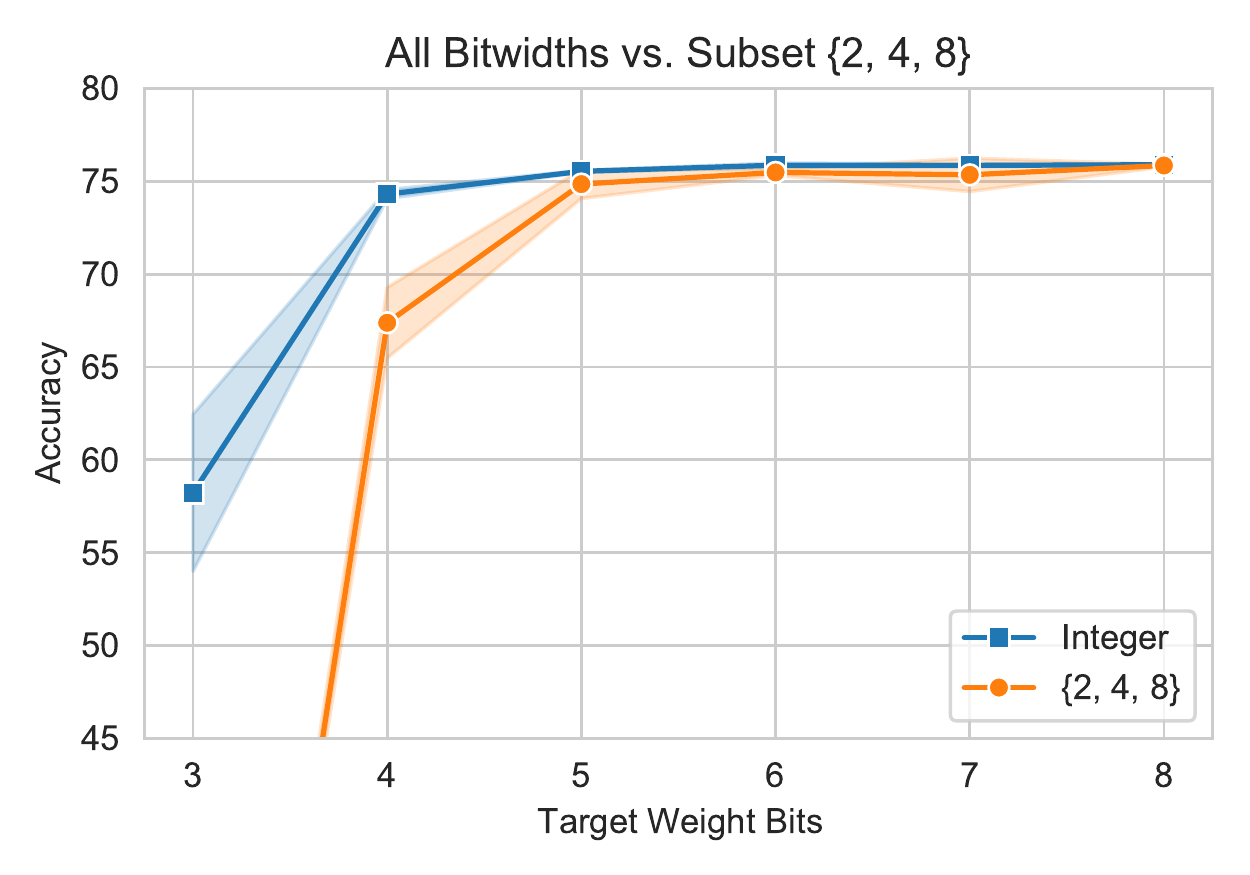}
    \caption{At high bitwidths, discrete and integer bitwidths retain comparable accuracy but at ultra-low precision constraining bitwidths to a discrete set performances far worse. }
    \label{fig:discrete_bitwidth}
\end{figure}

We test the impact on performance of using any integer bitwidth versus the subset $\{2, 4, 8\}$. We implement a restricted version of the post-training quantization method that rounds the continuously parameterized bitwidth to the closest of the three discrete values $\{2, 4, 8\}$ during training. As shown in Figure \ref{fig:discrete_bitwidth}, restricting the set of bitwidths used by the mixed-precision method does not affect accuracy at high average bitwidths, while at lower bitwidths the degradation is large. 

\paragraph{Observation 6: Mixed-precision activations constrained by the maximum feature map size have higher accuracy than uniform precision with the same maximum feature map size.} $\quad$

\begin{figure}
    \centering
    \includegraphics[width=\linewidth]{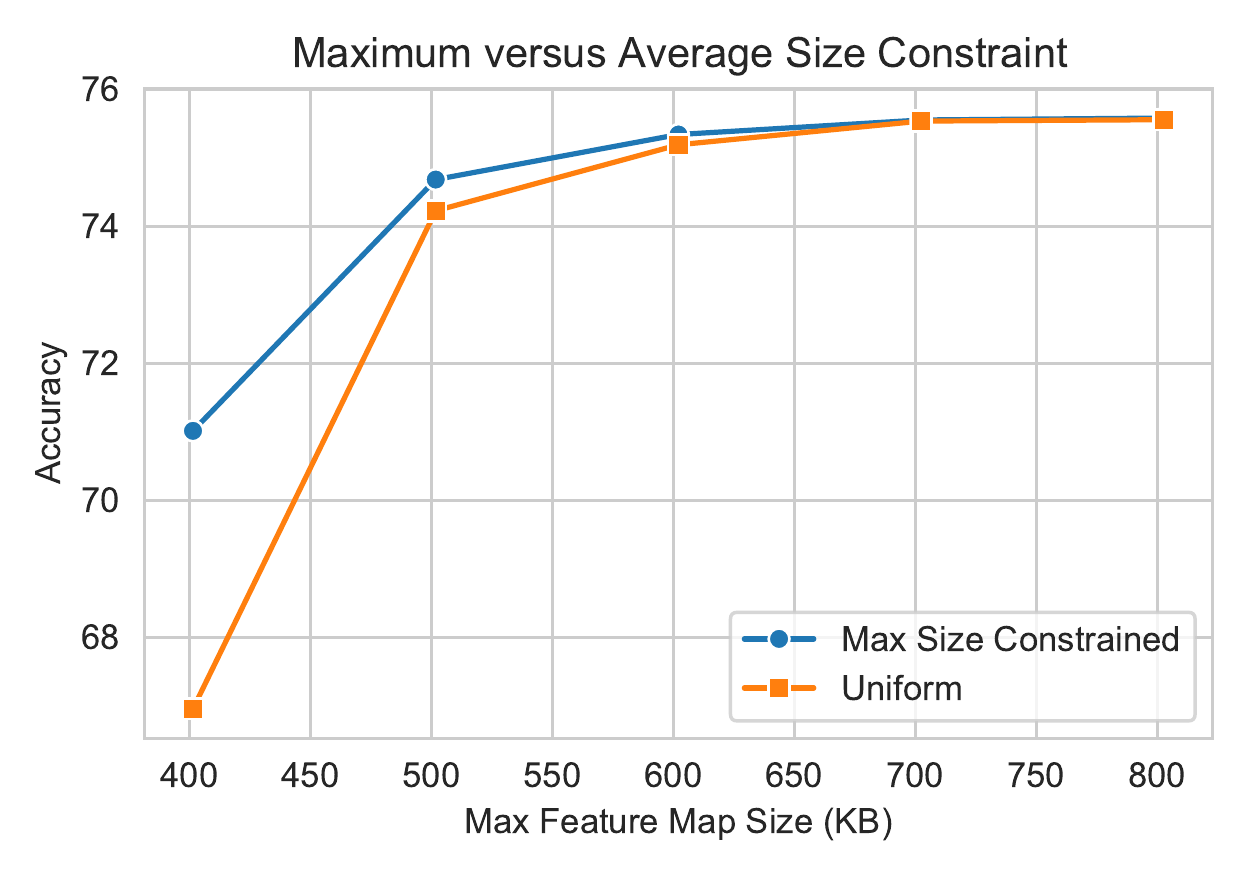}
    \caption{Mixed precision constrained by the maximum feature map size outperforms uniform precision with the same maximum feature map size.}
    \label{fig:max-act}
\end{figure}

We compare two settings: mixed-precision activations with a maximum feature map size, and uniform precision activations with the same maximum feature map size. For mixed-precision activations, we set the bitwidth of each layer so that all feature maps have the maximum feature map size, with a maximum bitwidth of 32. We show results in Figure \ref{fig:max-act}. For Resnet50, the model with $\sim$400KB maximum feature map size corresponds to a uniformly quantized model with 4 bit activations.   

\paragraph{Observation 7: Post-training quantization methods that use labeled data can perform equally well with pseudolabels.} $\quad$

\begin{figure}
    \centering
    \includegraphics[width=\linewidth]{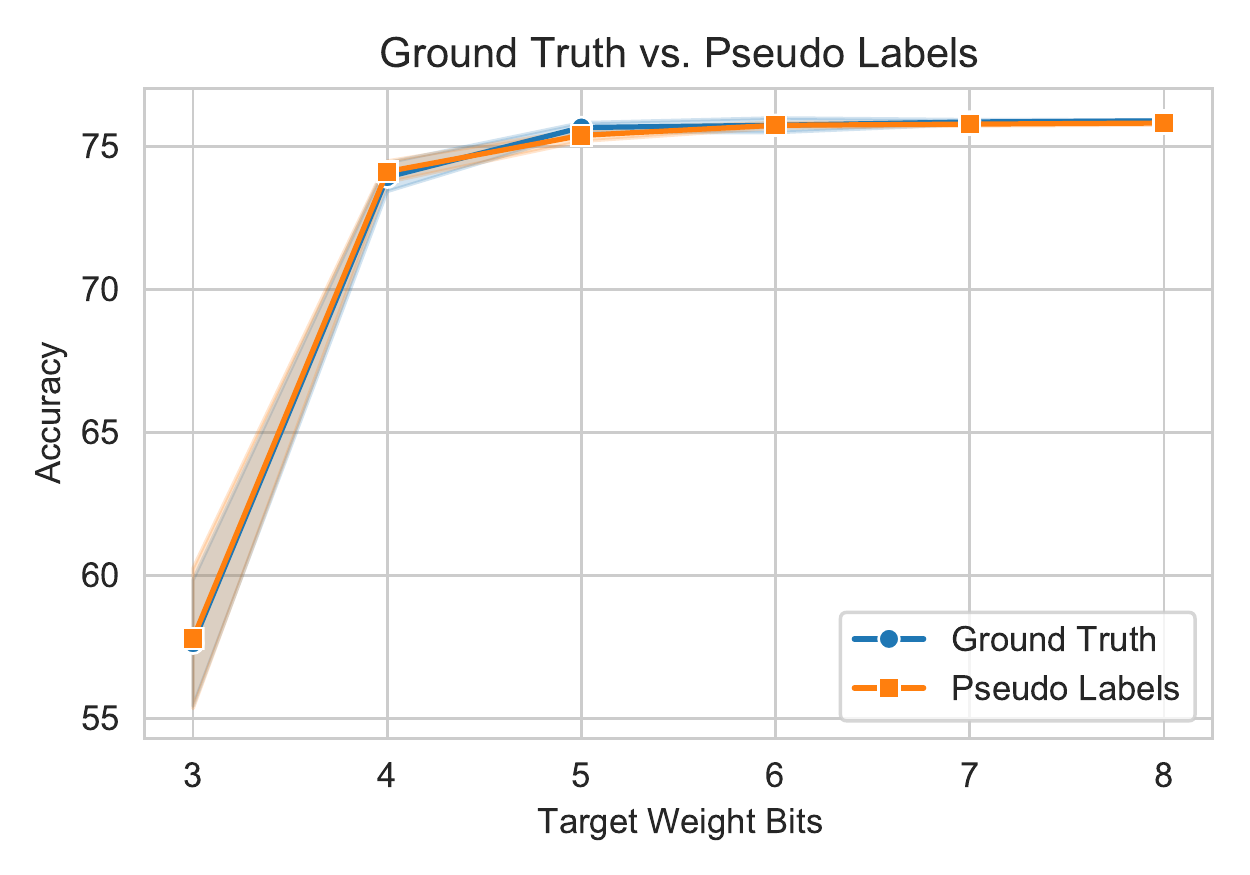}
    \caption{Post-training mixed-precision quantization can obtain similar accuracy even when ground truth labels are not available.}
    \label{fig:pseudolabels}
\end{figure}

While we use labeled training data for learning mixed-precision bit allocations, it is possible to generate labels using the floating point baseline, and train the model against those pseudolabels. In Figure \ref{fig:pseudolabels}, we show that the quantized model accuracy is close to identical with ground truth labels compared to pseudolabels. The accuracy of such approaches may be further improved by using soft pseudolabels as in knowledge distillation or self training.
This demonstrates that the data constraints of post-training quantization, e.g. whether using synthetic data or pseudolabels instead of ground truth training data, may be separated from the quality of quantization algorithms.

\paragraph{Observation 8: The floating point baseline accuracy of a model impacts quantized accuracy.} $\quad$

\begin{figure}
    \centering
        \includegraphics[width=\linewidth]{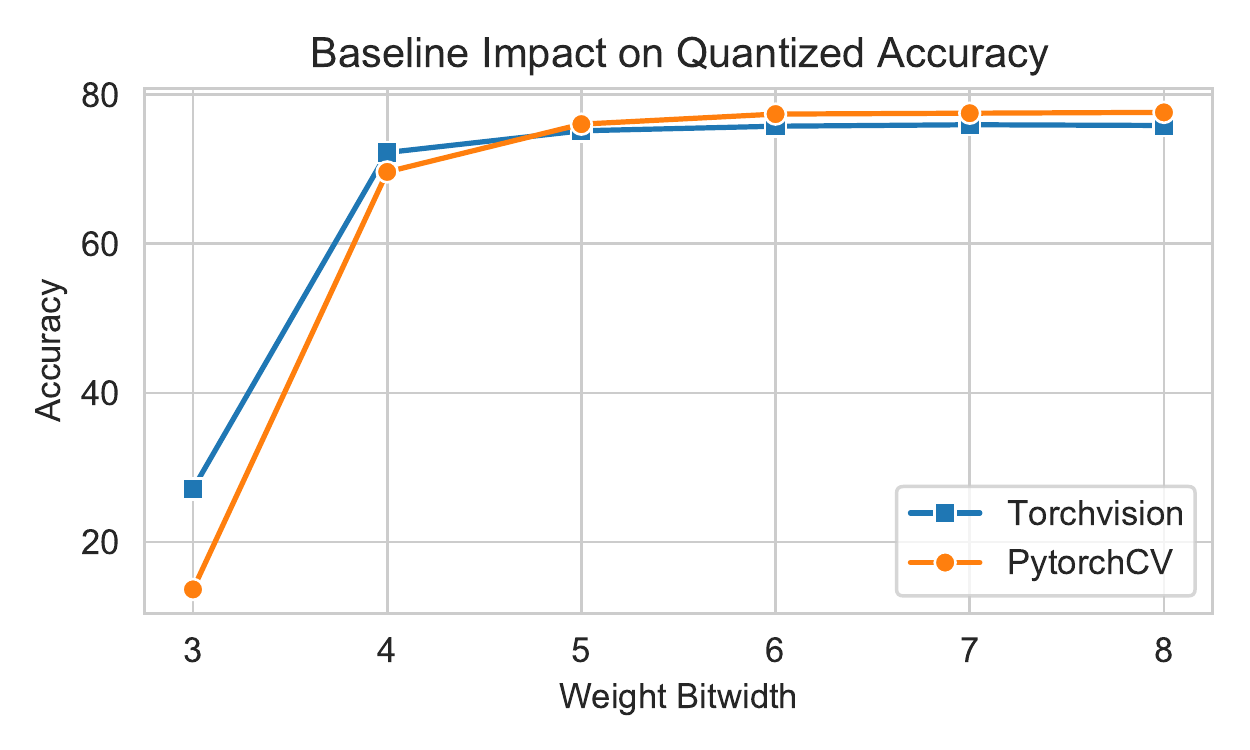}
    \caption{Floating point baseline affects results.}
    \label{fig:ptcv}
\end{figure}

We evaluate the accuracy obtained by the pytorchcv model (baseline accuracy 77.7\%) and torchvision model (baseline accuracy 76\%) with varying weight bitwidth in Figure \ref{fig:ptcv}. By virtue of the pretrained accuracy, the 5 bit quantized pytorchcv model outperforms all pytorch quantized models. However, at extremely low bitwidths, pretrained models with higher accuracy may not always perform better. 

\section{Discussion}
\label{sec:discussion}

\begin{table*}[t!]
\centering
\begin{tabular}{|p{0.2\linewidth}|p{0.75\linewidth}|}
 \hline
         &  Design Tradeoffs \\  \hline
Quantization Method &  Uniform $\vert$ Asymmetric  $\vert$ Per-channel weights, Per-tensor activations $\vert$ Static quantizer parameters $\vert$ Percentile clipping at 99.99th percentile for activations \\ \hline
Quantized Operations &  Full precision skip connection $\vert$ Full precision residual connection  $\vert$ Average pooling computed at high precision and requantized    \\ \hline
Mixed Precision & 8 bit weights $\vert$ Fixed activation bitwidth $\vert$ First and last layer weights at 8 bits, output activations of first and last layer at 8 bits\\ \hline
Resource Complexity &  320 examples $\vert$ Unlabeled training data $\vert$ ${<}1$ minute on GPU   \\ \hline
Pretrained Model &  Baseline accuracy: 76\% $\vert$ Quantized accuracy: Figure \ref{fig:highway_results}    \\ \hline
\end{tabular}
\caption{Quantization Card for Observation 3, Full Precision Skip Connections}
\label{tab:modelcard-obs2}
\end{table*}       

Given that the aforementioned confounding tradeoffs impact quantized model accuracy by ${>}1\%$, which may be as large as the improvements of state-of-the-art methods, we encourage all quantization papers to discuss and report their design tradeoffs through quantization cards. In Table \ref{tab:modelcard-obs2}, we present an example of a quantization card for the method used in Observation 3 which has full precision residual and skip connections, an often overlooked design tradeoff that the quantization card clearly highlights. In Appendix \ref{app:method}, we provide an example of a quantization card for the post-training mixed-precision method used in this work.

These tradeoffs can inform design decisions for developers deploying quantized models on hardware. Depending on primary bottlenecks (compute, memory bandwidth, or memory capacity), and acceleration opportunities (arithmetic with 2-4 bit integers), developers can use quantization cards to quickly determine whether the design tradeoffs made by the work are applicable for their use case. 

While these design tradeoffs are likely to manifest similarly for different methods, the magnitude of their impact may vary depending on the quantization method and precision setting. For example, symmetric quantization of weights is likely to increase the required bitwidth by 0.23 bits on average for methods that round weights to their nearest quantized value post-training \cite{banner:fourbit}, but may have a smaller impact for methods that do not round to nearest \cite{wang:bitsplit, nagel:adaround, hubara:adaquant} or perform quantization-aware training \cite{esser:lsq, Uhlich:learn_bw}. This emphasizes the importance of discussing the interactions between design tradeoffs and novel methods; if throughput on some hardware is substantially improved by using signed arithmetic, then methods that are less impacted by symmetric quantizers are preferable. 

We leave the creation of a benchmark to future work, after the design scenarios for quantization are better established.
Such a benchmark might include different resource constraints and quantization implementations, and allow developers to supply quantized model weights and quantizer parameters for the design setting of their choice. 
This will facilitate clearer evaluation between different approaches to quantization.

\section{Conclusion}
In this work, we demonstrate that novel quantization papers evaluate their results subject to design tradeoffs that are independent of the proposed method, though these tradeoffs can impact quantized network accuracy as much as the methods themselves. These include a variety of design decisions, some even seemingly innocuous, such as whether to use asymmetric/symmetric quantizers, per-channel vs. per-tensor quantization, quantized or full precision skip connections, the set of allowed bitwidths for mixed-precision quantization, the number of data points used for post-training quantization, and the baseline accuracy of the pretrained model. 
Given these results, we encourage novel quantization papers to explicitly report the design decisions of their work, and suggest a framework for reporting and evaluating these tradeoffs through ``quantization cards.''

\section*{Acknowledgment}
We would like to thank many members of the Luminous Computing team, including Rohun Saxena, for their helpful discussions and suggestions.

{\small
\bibliographystyle{ieee_fullname}
\bibliography{bibliography}
}

\newpage
\appendices
\section{Learning Bitwidths for Post-Training Mixed-Precision Quantization}
\label{app:method}

We describe the method for learning bitwdiths for mixed-precision post-training quantization formally, adapted from \cite{Uhlich:learn_bw}. 
We let $B_w^{(l)}$ and $B_a^{(l)}$ denote the number of bits used for weights and activations in layer $(l)$, respectively, and use $\bfB = \{B_w^{(l)}, B_a^{(l)}\}$ to denote a vector of all of the parameterized bitwidths in the network. Let $B_{w_{avg}}(\bfB), B_{a_{avg}}(\bfB)$ denote functions that compute the average number of bits used for weights and activations across the network, 
weighted by the the size of each weight matrix or intermediate activation map, 
and $B_{w_{max}}$ and $B_{a_{max}}$  denote the target maximum average number of weight and activation bits. We parameterize the bitwidths continuously, but round them to the nearest integer during the forward pass, allowing gradients to flow backwards using the Straight-through-Estimator (STE) for all rounding \cite{bengio:ste}. Then, our method requires solving the optimization problem
\begin{equation}
\begin{aligned}
\minimize_{\bfB} \quad  &-\mathbb{E}_{p(\bfx, y)}\left[\log {p}_m(y | \bfx; \theta, \bfB)\right]\\
  & + \lambda_1 \textrm{ max}\left(B_{w_{avg}}(\bfB) - B_{w_{max}}, 0\right)\\
  & + \lambda_2 \textrm{ max}\left(B_{a_{avg}}(\bfB) - B_{a_{max}}, 0\right)\\
\end{aligned}
\end{equation}

We note several key differences from \cite{Uhlich:learn_bw}. First, we parameterize directly with respect to the number of bits. Unlike \cite{Uhlich:learn_bw}, we observe no difference in empirical performance between parameterizing with respect to the bitwidth and the scale, and find that parameterizing with respect to the bitwidth is more natural for per-layer bitwidth allocation when using per-channel quantization parameters. Second, the optimization problem is post training, and is consequently only with respect to $\bfB$, not the parameters $\theta$. Third, because the number of samples is limited, we only train minimum and maximum clipping range for per-tensor quantization of activations, and not for per-channel weight quantizers. Training both may require hyperparameter tuning of learning rates for different parameter groups, as in \cite{hubara:adaquant}.
We evaluate the model with the highest exponential moving average train accuracy that meets the constraints during training because the discrete nature of the optimization problem means that the model does not always meet the constraints at the final training step. 

We show a quantization card for this method in Table \ref{tab:modelcard-mp}.

\begin{table*}[t!]
\centering
\begin{tabular}{|p{0.2\linewidth}|p{0.75\linewidth}|}
 \hline
         &  Design Tradeoffs \\  \hline
Quantization Method &  Uniform $\vert$ Asymmetric  $\vert$ Per-channel weights, Per-tensor activations $\vert$ Static quantizer parameters $\vert$ Learned activation quantizer min/max \\ \hline
Quantized Operations &  Quantized skip connection $\vert$ Quantized residual connection  $\vert$ Average pooling computed at high precision and requantized    \\ \hline
Mixed Precision & Mixed precision weights $\vert$ Mixed precision activations $\vert$ All layers quantized $\vert$ Any integer bitwidth \\ \hline
Resource Complexity &  1024 examples $\vert$ Labeled training data $\vert$ $4\%$ of one epoch train time   \\ \hline
Pretrained Model &  Baseline accuracy: 76\% $\vert$ Quantized accuracy: Table \ref{tab:first+last}    \\ \hline
\end{tabular}
\caption{Quantization Card for Mixed-Precision Post-Training Quantization}
\label{tab:modelcard-mp}
\end{table*}       

\section{Experimental Details}
\label{app:details}

Unless otherwise stated, all experiments use uniform, asymmetric, static quantization, with per-tensor quantization of activations and per-channel quantization of weights. Skip and residual connections are quantized. Average pooling is performed with high precision accumulators and the outputs are requantized. We use the torchvision model with 76\% baseline validation accuracy. 

When quantizing activations to low precision uniformly (Figure \ref{fig:unif-symmetric}, Figure \ref{fig:highway_results}), we use percentile clipping of activations at the 99.99th percentile, as in \cite{li:percentileclip, mckinstry:percentile}. The first and last layer are kept at 8 bits, and all weights are 8 bits. 320 unlabeled examples are used for percentile clipping. Experiments in Figure \ref{fig:unif-symmetric} quantize skip connections but not residual connections, while experiments in Figure \ref{fig:highway_results} vary the strategy for quantizing residual and skip connections. 

For low precision weight quantization when using uniform precision (Figure \ref{fig:ptcv}, Figure \ref{fig:channel}), we use min/max calibration of activations and weights. The first and last layer are kept at 8 bits, and all activations are 8 bits. 1024 unlabeled examples are used for calibration.

Experiments with a maximum feature map constraint (Figure \ref{fig:max-act}) keep the last layer at 8 bits, but quantize the first layer to low precision since the outputs of the first layer have the maximum feature map size. We use percentile clipping for activations.  

For learned mixed precision of weights only (Figure \ref{fig:weight-symmetric}, Figure \ref{fig:discrete_bitwidth}), we use min/max calibration of activations and allow the bitwidths of all layers, including the first and last, to be any integer. Compression ratios include the first and last layer. All activations are 8 bits. For the bitwidth subset in Figure \ref{fig:discrete_bitwidth}, 2, 4, and 8 bits are allowed. Mixed precision allocations are trained on 1024 labeled data examples for 1500 steps with a batch size of 32, which takes ${<}$20 minutes on a NVIDIA V100 GPU. Parameters are trained with the Adam optimizer \cite{kingma:adam} with a learning rate of 0.01.

When learned mixed precision of both activations and weights (Table \ref{tab:first+last}), we use min/max calibration of weights, and allow the min/max for activations to be learned jointly with bit allocations. All integer bitwidths are allowed, and compression ratios include the first and last layer. For experiments when the compression ratio includes a 8-bit first and last layer, the bitwidth of the first and last layer is fixed. We use the same training setup as for mixed precision weights. 

\end{document}